\documentclass[10pt,twocolumn,letterpaper]{article}

\usepackage{iccv}
\usepackage{times}
\usepackage{epsfig}
\usepackage{graphicx}
\usepackage{amsmath}
\usepackage{amssymb}
\usepackage{multirow}
\usepackage{enumitem}
\usepackage{subcaption}
\usepackage[ruled,vlined]{algorithm2e}
\usepackage{textcomp}
\usepackage{pifont}

\usepackage[breaklinks=true,bookmarks=false]{hyperref}

\iccvfinalcopy 

\def\iccvPaperID{2499} 

\usepackage{amssymb}
\usepackage{pifont}
\newcommand{\cmark}{\ding{51}}%
\newcommand{\xmark}{\ding{55}}%

\ificcvfinal\fi

\begin{document}

\title{Exploring Simple 3D Multi-Object Tracking for Autonomous Driving}

\author{
  Chenxu Luo${^{1,2}}$ \quad Xiaodong Yang${^1}$\thanks{Correspondence to \texttt{xiaodong@qcraft.ai}} \quad Alan Yuille${^2}$\\
  $^1$QCraft \quad $^2$Johns Hopkins University \\
}

\maketitle
\ificcvfinal\thispagestyle{empty}\fi

\begin{abstract}
3D multi-object tracking in LiDAR point clouds is a key ingredient for self-driving vehicles. Existing methods are predominantly based on the tracking-by-detection pipeline and inevitably require a heuristic matching step for the detection association. In this paper, we present SimTrack to simplify the hand-crafted tracking paradigm by proposing an end-to-end trainable model for joint detection and tracking from raw point clouds. Our key design is to predict the first-appear location of each object in a given snippet to get the tracking identity and then update the location based on motion estimation. In the inference, the heuristic matching step can be completely waived by a simple read-off operation. SimTrack integrates the tracked object association, newborn object detection, and dead track killing in a single unified model. We conduct extensive evaluations on two large-scale datasets: nuScenes and Waymo Open Dataset. Experimental results reveal that our simple approach compares favorably with the state-of-the-art methods while ruling out the heuristic matching rules.
\end{abstract}

\section{Introduction}

3D multi-object tracking is a crucial component in an autonomous driving system as it provides pivotal information to facilitate various onboard modules ranging from perception, prediction to planning. LiDAR is the most commonly used sensor that a self-driving vehicle relies on to perceive its surroundings. Thus, tracking in LiDAR point clouds has been attracting increasing interests with the rapid development of self-driving vehicles in recent years. 

Multi-object tracking is a long-standing task in computer vision and has been extensively studied in image sequence domain. Arguably, the tracking-by-detection is the most popular tracking paradigm, which first detects objects for each frame and then associates them across frames. These methods have shown promising results and benefited from huge progress in image object detection. They usually formulate the association step as a bipartite matching problem. Most existing works therefore focus on better defining the affinity matrix between tracked objects and new detections. In the matching criteria design, the motion~\cite{sort} and appearance~\cite{zhan2020simple} are widely adopted as the association cues. 

\begin{figure}[t]
\begin{center}
\includegraphics[width=\linewidth]{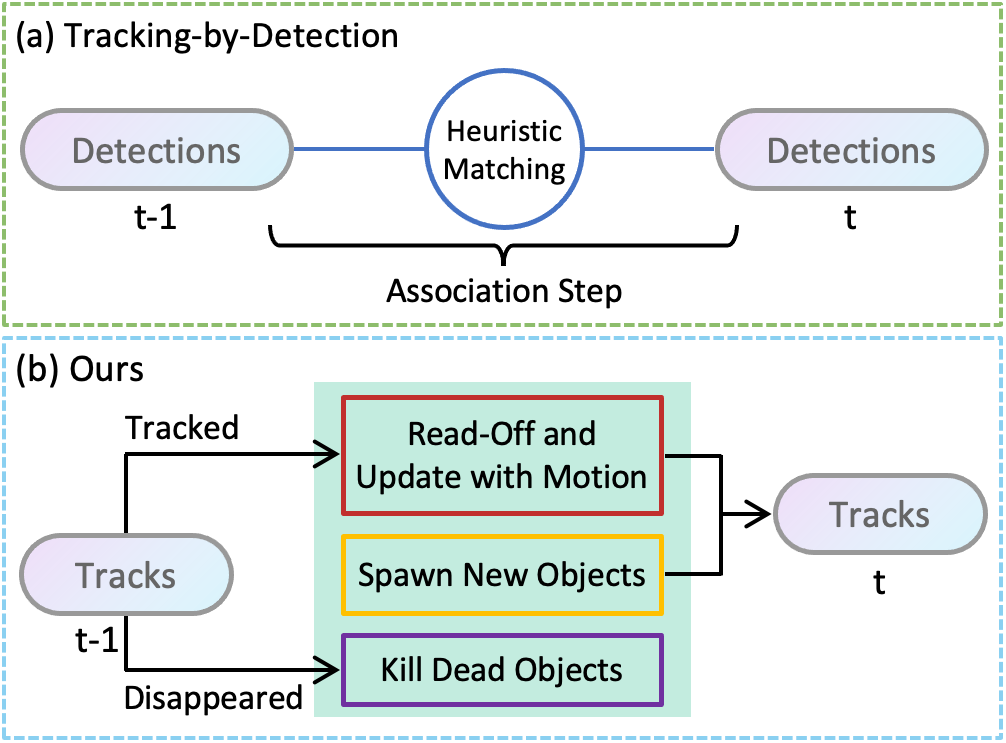}
\end{center}
   \vspace{-12pt}
   \caption{An overview of the tracking-by-detection pipeline and our approach. (a) performs 3D object detection in each point cloud and then match detected objects through an association step, which involves complex heuristic rules. (b) reads-off tracking identities and updates object locations using estimated motion, and at the same time manages new-born and dead tracks. Our model handles the three cases in a single forward pass without requiring heuristic matching.}
   \vspace{-1pt}
\label{fig:teaser}
\end{figure}


For 3D multi-object tracking with LiDAR, the tracking-by-detection pipeline also plays the dominant role~\cite{chiu2020probabilistic,weng20203d}. Accordingly, in order to obtain the final tracking result, current methods inevitably require a heuristic matching step to link detected objects over time in a separate stage. There exists numerous hand-crafted rules when performing such a step. As compared in the supplementary material, different matching criteria and corresponding threshold for each specific object class substantially impact the final tracking performance. This also happens in the track life management, which is used to handle new-born objects and dead tracks. It is a common practice for these methods to initialize a track only when an object continuously presents for a certain number of frames in order to filter out false detections, and keep disappeared objects for several frames to tackle occlusion.
Unfortunately, all these heuristic rules are not trainable and highly depend on their hyper-parameters that demand huge efforts to tune. What is worse, such rules and hyper-parameters are often data and model dependent, making it hard to generalize and laborious to re-tune when applying to new scenarios. 



The main reason for the requirement of an additional heuristic matching step is the lack of connection between frames in conducting object detection.
Recently, some methods~\cite{peng2020chained,yin2020center} estimate velocity or predict the location of an object in consecutive frames to provide such a connection across frames. However, they merely treat the forecasted detections as a bridge for object matching instead of using them as the final tracking output. Moreover, they only take into account the location relationship of objects between frames, without modeling the confidence for the association. So the confidence score only reflects the detection confidence in a single frame. As a result, these methods are often prone to spurious detections and have to manually decide the number of frames to keep for the purpose of dealing with occluded objects. Another issue lies in how to process newborn objects in an online tracking system. Existing methods~\cite{bergmann2019tracking, peng2020chained} re-detect all objects in the current frame so that matching is still in need to differentiate the newborn objects from the tracked ones. 


In light of the above observation, we present \textbf{SimTrack}: a simple model for 3D multi-object tracking in point clouds. We simplify the existing hand-crafted tracking algorithms without requiring the heuristic matching step. Our approach is flexible to build upon the commonly used pillar or voxel based 3D object detection networks~\cite{lang2019pointpillars,zhou2018voxelnet}. We propose a novel hybrid-time centerness map, which represents objects by their first-appear locations within the given input period. Based on this map, we can directly link current detections to previous tracked objects without the need for additional matching.
Since this map formulates the detection and association of an object at the same time, our model is able to inherently provide the association confidence between frames. 
Also, we introduce a motion updating branch to estimate the motion of tracked objects in order to update from their first-appear locations to the current positions.
For the new-born objects and dead tracks, they can be determined simply by confidence thresholding as regular detections on the same map, thus removing the manual track life management too.
As illustrated in Figure~\ref{fig:teaser}, our model eliminates the heurstic matching step, and unifies the tracked object linking, new-born object detection as well as dead track killing in a single forward pass. 


To our knowledge, this work provides the first learning paradigm that is able to get rid of the heuristic matching step for 3D multi-object tracking in point clouds, and therefore, remarkably simplifies the whole tracking system. We introduce a novel end-to-end trainable model for joint detection and tracking through a hybrid-time centerness map and a motion updating branch. Experimental results reveal that this simple approach compares favorably to the existing methods. Our code and model will be made available at \url{https://github.com/qcraftai/simtrack}. 


\begin{figure*}[t]
\begin{center}
\includegraphics[width=\linewidth]{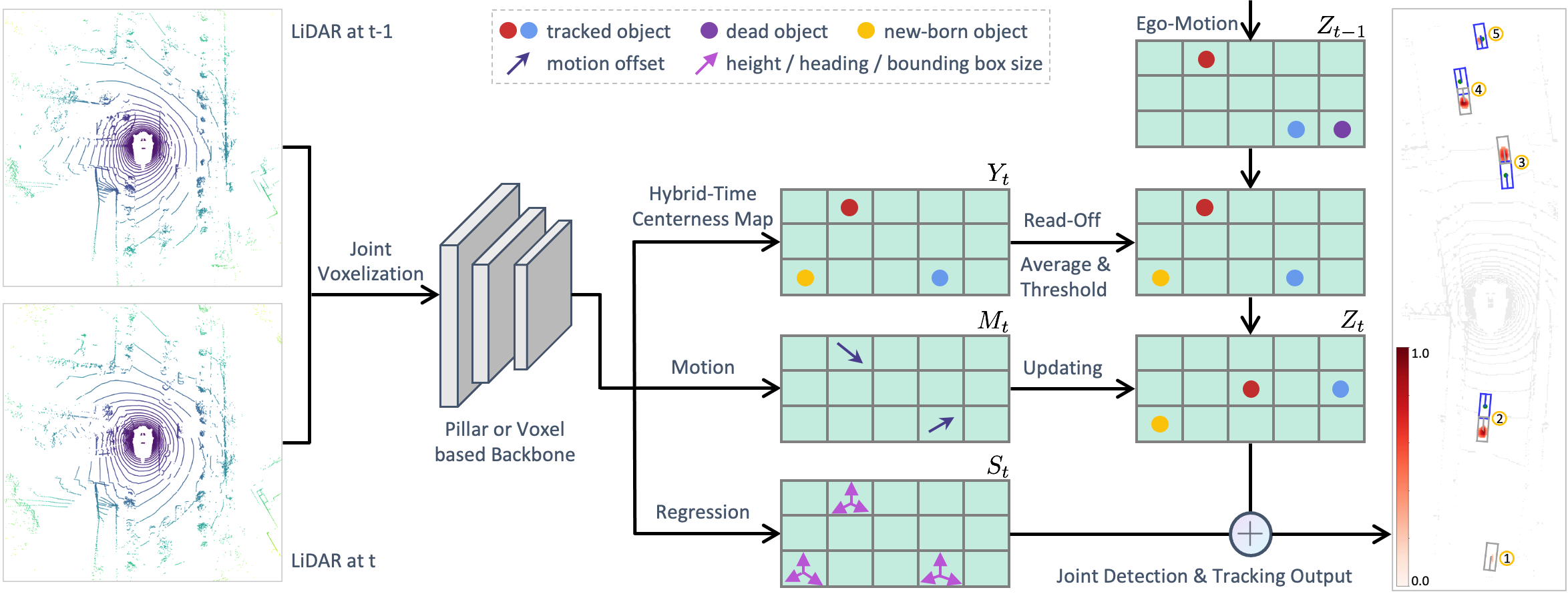}
\end{center}
\vspace{-10pt}
   \caption{A schematic overview of SimTrack. Our model consists of a hybrid-time centerness map branch that detects the first-appear location of each object in the input snippet, a motion updating branch to predict the motion of an object within the period, and a regression branch to estimate other object properties. During the inference, we first transform the previous updated centerness map $Z_{t-1}$ to the current coordinate system through ego-motion, and average it with the current hybrid-time centerness map $Y_t$ which is next thresholded to remove the dead objects, and then read-off the tracking identities that share the same cells on $Y_t$ and $Z_{t-1}$. After that, we update the tracked objects to their current locations using the predicted motion to obtain $Z_{t}$. We show a zoom-in area of the point cloud to illustrate the detection and tracking output, where the gray and blue boxes are detected objects at previous and current sweeps. ID (1) is a dead object with low confidence. ID (2-4) are tracked objects that are correctly localized from peaks in the confidence heatmap, and their current locations are accurately updated by the predicted motion. ID (5) is a new-born object.}   
\label{fig:overview}
\end{figure*}

\section{Related Work}

\textbf{2D Multi-Object Tracking.} 
With continuous progress in image object detection~\cite{ren2015faster,ren2020instance,ren2020ufo,tian2019fcos}, most methods follow the tracking-by-detection pipeline that first detects objects for individual frames and then associates two sets of detections over time. We can categorize the association step into two primary groups: motion based and appearance based. The motion based methods utilize temporal modeling~\cite{prernn} to update detections and realize matching with a distance or intersection over union metric (IOU). Kalman filter~\cite{kalman1960new} is widely adopted by the methods of this group for state estimation~\cite{sort}. Some works predict location offset to facilitate motion modeling~\cite{feichtenhofer2017detect,peng2020chained,zhou2020tracking}. The appearance based methods instead consider visual appearance affinity for the same object in a sequence. Most of them apply re-identification~\cite{zheng2019cvpr,zou2020eccv} to drive appearance feature learning to establish identity correspondence~\cite{ lu2020retinatrack,zhan2020simple}. 

Unlike the motion based tracking-by-detection methods that treat the predicted boxes only as a proxy for matching, the tracking-by-regression paradigm performs tracking by directly regressing previous locations to new positions in current frame. In~\cite{bergmann2019tracking} Trackor starts from the past location of each object and applies the region-pooled object detection features from current frame to obtain the updated location. It depends on an additional object detector to deal with new-born objects, and has to update multiple regions of interest and separate tracked objects from new detected objects via some heuristic rules. In contrast, our approach can directly produce the tracked and new-born objects in a single forward pass with no heuristic post processing. 

Recently, inspired by the success of Transformer~\cite{vaswani2017attention}, several methods are developed to conduct joint detection and multi-object tracking through attention operations.
TrackFormer~\cite{meinhardt2021trackformer} employs track query embeddings to follow object location changing over time in an auto-regressive way and adopts object queries as in~\cite{detr} to deal with new-born objects.
TransTrack~\cite{sun2020transtrack} adopts the query-key mechanism to detect objects in current frames and associate them between frames by the learned object queries.

\textbf{3D Multi-Object Tracking.} In this field, the dominant methods exploit the tracking-by-detection pipeline. Due to the lack of appearance and texture cues in point clouds, the LiDAR based tracking models hinge on motion for association. AB3DMOT~\cite{weng20203d} extends Kalman filter to 3D for motion state estimation. CenterPoint~\cite{yin2020center} estimates the velocity of each object by adding a velocity regression head.
FaF~\cite{luo2018fast} associates objects through prediction. 
PnPNet~\cite{liang2020pnpnet} learns affinity matrix between objects using 3D features and trajectories. Chiu et al.~\cite{chiu2021} incorporate appearance feature distance from cameras into the distance metric to enhance association. Most of them require a bipartite matching step, either using the Hungarian algorithm or a greedy matching algorithm, to acquire the final tracking output. 

\textbf{Detection and Motion.} 3D object detection provides the fundamental tools for 3D multi-object tracking. In~\cite{zhou2018voxelnet} VoxelNet applies 3D convolutions on the voxel features extracted by~\cite{qi2017pointnet}. SECOND~\cite{yan2018second} enhances efficiency by using sparse 3D convolutions. CBGS~\cite{zhu2019class} improves accuracy by class-balanced grouping and sampling. PointPillars~\cite{lang2019pointpillars} is developed to collapse height dimension and employ 2D convolutions to achieve better efficiency.
Meanwhile, some recent methods~\cite{pillar-motion,wu2020motionnet} demonstrate promising results for point cloud pillar motion estimation using self-supervision or proxy motion supervision derived from tracking.


\section{Method}
As illustrated in Figure~\ref{fig:overview}, SimTrack unifies the tracked object linking, new-born object detection and dead object removal in a single end-to-end trainable model. Our key design to rule out the heuristic matching step and achieve the desired simplified tracking is based on the proposed hybrid-time centerness map and motion updating branch. 

\subsection{Preliminary}
Our approach exploits the center based representation for 3D objects. Thanks to the innate connection between detection and tracking under this representation, tracks can be described as paths formed by points in space and time. Here we briefly review the center based 3D object detection. 
Given a raw point cloud, we first voxelize it into regular grids using either pillars~\cite{lang2019pointpillars} or voxels~\cite{zhou2018voxelnet}. We extract the feature of each pillar or voxel by a small PointNet~\cite{qi2017pointnet}. After that the standard 2D or 3D convolutions are used to compute the features in bird's eye view (BEV). As for the detection head, we represent each object by its center location on the centerness map similar to~\cite{yin2020center}. For training, a 2D Gaussian heatmap around the center of each object is created to form the target centerness map. All detection output including the centerness map, local offset, object size and heading can be generated by the detection head.


\subsection{Overview}
Let $P^t = \{(x, y, z, r)_i\}$ indicate an orderless point cloud consisting of the measurements of coordinates $(x, y, z)$ and reflectance $r$ at time $t$. 
Our model takes a snippet of point clouds as input. For simplicity, we directly combine multiple point clouds by transforming the past sweeps to the current coordinate system through ego-motion compensation. As a common practice, we also add a relative timestamp to each point such that a point is represented as $(x, y, z, r, \Delta_t)$, where $\Delta_t$ is the relative timestamp to the current sweep. After the voxelization and feature extraction, our detection head makes use of the centerness map to detect the first-appear location of an object in the input snippet and estimates the object motion within the period. In the inference, we simply read-off the tracking identity of an object from previous centerness map and then use the predicted motion to update the object to its current location.

\subsection{Joint Detection and Tracking}

As discussed above, in order to eliminate the heuristic matching and manual track life management, we propose to perform joint detection and tracking in a simplified model through the combination of a hybrid-time centerness map and a motion updating branch. 

\textbf{Hybrid-Time Centerness Map.} With the intention to provide linkage with previous detections and detect new-born objects simultaneously, we propose a hybrid-time centerness map. Specifically, our model takes two consecutive LiDAR sweeps at $t-1$ and $t$ as input. For the target centerness map, we represent each object by its center location, where this object first appears in the input sequence. Suppose that the ground-truth object locations at frame $t-1$ and $t$ are respectively $\{d_i^{t-1}\}_{i=1,\cdots,n_{t-1}}$ and $\{d_i^{t}\}_{i=1,\cdots,n_{t}}$. Our target assigning strategy is defined as follows. 

\begin{itemize}[noitemsep,topsep=0pt]
	\item For a tracked object that presents in both $t-1$ and $t$ frames, denoted as $d_{i}^{t-1}$ and $d_{j}^{t}$, we create its target heatmap at $d_{i}^{t-1}$, which is the first-appear location of this object in the input sequence.  
	\item For a dead object that only shows up in the first frame $t-1$ but disappears in the second frame $t$, we treat it as a negative example and do not assign any target heatmap for this object. 
	\item For a new-born object that only presents in the second frame $t$, we create its target heatmap at $d_{i}^{t}$. 
\end{itemize}

In this manner, for a tracked object, we can directly link with its previous detection by reading-off the identity at same location of the updated centerness map (see details bellow) from the previous timestamp. For the dead objects, they can be simply removed by thresholding the confidence scores. And for the new-born objects, we perform regular detection on the same hybrid-time centerness map. Therefore, our hybrid-time centerness map builds the foundation to merge tracked object association, dead object removal, and new-born object detection in a single unified model. Furthermore, we can utilize the confidence scores obtained from this hybrid-time centerness map to imply both detection confidence (i.e., the probability of an object existing in current timestamp) and association confidence (i.e., the probability of an object linking to its previous location).

\textbf{Motion Updating Branch.} As aforementioned, we link each tracked object to its previous location on the hybrid-time centerness map to establish identity correspondence. However, to achieve an online tracking system, we need to further obtain the current location of the object. We thus introduce a motion updating branch to estimate the offset of an object between two sweeps. In practice, for each object, at its center location in the first frame, we regress the offset to its current location: $(\Delta_u, \Delta_v) = (u_{t}-u_{t-1}, v_{t}-v_{t-1})$, where $(u,v)$ is the object center coordinates. We then take advantage of this motion field to transform a hybrid-time centerness map into an updated centerness map.

We note that some previous methods such as CenterPoint~\cite{yin2020center} also estimates object velocity. However, the main difference is that they only regard motion as an assistant to conduct matching. They use motion to propagate the detections of current frame and use them to match with the detections from previous frame. In other words, the propagated boxes only serve as a bridge for matching the detected objects across frames, instead of being used as the final tracking result. SimTrack however shows that our hybrid-time detection and motion estimation can be combined to produce the tracking output without the need of heuristic matching. 
Moreover, during the inference, CenterPoint requires to manually tune a class-specific distance threshold to determine whether a motion-derived box can be matched with a detected box. In comparison, our model remarkably simplifies tracking pipeline through a single forward pass to obtain both detections and correspondences.  



\textbf{Other Regression Branches.} In addition to the motion, we regress other 3D object properties including height $z$, bounding box size $(w, l, h)$, and heading in the format of $(\sin\theta, \cos\theta)$ where $\theta$ is the yaw angle of the bounding box.

\textbf{Loss Functions.} When training the hybrid-time centerness map, we adopt the focal loss similar to~\cite{yin2020center, zhou2019objects}:
\begin{equation}
    \hspace{-2mm}\mathcal{L}_\text{cen}=\frac{-1}{N}\sum\limits_{c,d_i} \begin{cases}
    (1-Y_{c,d_i})^\alpha\log(Y_{c,d_i}), &\hspace{-2mm}\text{if } \Tilde{Y}_{c,d_i} = 1 \\
    (1-\Tilde{Y}_{c,d_i})^\beta(Y_{c,d_i})^\alpha\\\hspace{4mm}\log(1-Y_{c,d_i}), &\hspace{-2mm}\text{otherwise}
    \end{cases}
\end{equation}
where $\Tilde{Y}$ and $Y$ indicate the target and predicted hybrid-time centerness maps, $N$ denotes the number of objects, $\alpha$ and $\beta$ are the hyper-parameters of focal loss~\cite{lin2017focal}. For the motion updating branch, we enforce the standard $\ell_1$ loss:
\begin{equation}
    \mathcal{L}_\text{mot}=\frac{1}{N}\sum\limits_{i=1}^{N}|\Tilde{M}_{d_i}-M_{d_i}|,
\end{equation}
where $\Tilde{M}$ denotes the ground truth motion map, and $M$ is the predicted motion map. Similarly, we also make use of the standard $\ell_1$ loss for other regression branches: 
\begin{equation}
    \mathcal{L}_\text{reg}=\frac{1}{N}\sum\limits_{i=1}^{N}|\Tilde{S}_{d_i}-S_{d_i}|,
\end{equation}
where $\Tilde{S}$ and $S$ represent other ground truth and predicted regression maps for object height, size and heading. We only compute these losses at the center locations $d_i$ on the corresponding maps. In summary, the total objective is a weighted sum of the three loss functions: 
\begin{equation}
    \mathcal{L}_\text{total} = \omega_\text{cen}\mathcal{L}_{\text{cen}} +  \omega_\text{mot}\mathcal{L}_{\text{mot}} + \omega_\text{reg}\mathcal{L}_{\text{reg}},
    \label{eq:total}
\end{equation}
where $\omega_\text{cen}$, $\omega_\text{mot}$ and $\omega_\text{reg}$ are the balancing coefficients to control the importance of three loss terms. 

\textbf{Backbone Network.} SimTrack is flexible to be built on various backbones. In the experiments, we mainly use PointPillars~\cite{lang2019pointpillars} as the pillar based backbone due to its computational efficiency for onboard deployment. To compare with other methods, we also evaluate with the more accurate and larger VoxelNet~\cite{zhu2019class} as the voxel based backbone. 

\textbf{Online Inference.} During the inference, the updated centerness map $Z$ records the tracking identity, center location, and confidence score of each object. And the tracking identity is placed at the object center location.
For the initial frame in a sequence, our approach takes only one sweep as input and performs detection to initialize the updated centereness map $Z_0$. For the later frames, the model takes current sweep and one previous sweep as input. All point clouds are transformed to current vehicle coordinate system using ego-motion. 

In comparison to the existing methods that depend on a heuristic matching step, SimTrack uses a simple read-off to establish the association. As illustrated in Figure~\ref{fig:overview}, at time $t$ we first transform the updated centerness map $Z_{t-1}$ of previous timestamp to the current coordinate system using ego-motion. We then average $Z_{t-1}$ with the current hybird-time centerness map $Y_t$. For each object center, if there is an existing tracking identity at the same location on $Z_{t-1}$, the object is regarded as a tracked object and reads-off this tracking identity. We initialize a new track for each of the rest object centers. In our approach, there is no need to specifically handle the dead objects as they can be naturally discarded when thresholding $Y_t$. Afterwards, we update $Y_t$ to $Z_t$ by using the predicted motion map $M_t$ to obtain the current locations of tracked objects. We summarize the inference outline of our approach in Algorithm~\ref{algo}.

\begin{algorithm}[t]
\textbf{Input:} a sequence of point clouds $P_0, P_1, \cdots$\\
\textbf{Output:} joint detection and tracking output\\
\SetAlgoLined
 \For{$t = 0, 1, \cdots$}{
  \eIf{$t == 0$}{
   $Y_0, M_0, S_0 \leftarrow \text{Network}(P_0)$\\
   threshold and NMS on $Y_0$\\
   initialize tracking identities on $Y_0$\\
   $Z_0 \leftarrow Y_0$\\
   }{
   transform $Z_{t-1}$ by ego-motion\\
   $Y_t, M_t, S_t \leftarrow \text{Network}(P_t, P_{t-1})$\\
   $Y_t \leftarrow (Y_t + Z_{t-1}) / 2$\\
   threshold and NMS on $Y_t$\\
   read-off tracking identities from $Z_{t-1}$ to $Y_t$\\
   initialize new-born tracking identities on $Y_t$\\
   $Z_t \leftarrow \text{Update}(Y_t, M_t)$\\
  }
 }
 \caption{Online Inference of SimTrack}
 \label{algo}
\end{algorithm}

\section{Experiments}
In this section, we first describe our experimental setup including datasets, evaluation metrics and implementation details. A variety of ablation studies and related analysis are then provided for in-depth understanding of different design choices in our approach. We report extensive comparisons with the state-of-the-art methods on the two benchmarks.  

\subsection{Datasets}
We extensively evaluate our proposed approach on the two large-scale autonomous driving datasets: nuScenes~\cite{caesar2020nuscenes} and Waymo Open Dataset~\cite{sun2020scalability}. \textbf{nuScenes} contains 1000 scenes, each of which is around 20 seconds, and the point clouds are captured by a 32-beam LiDAR. This dataset is split to 700, 150 and 150 scenes for training, validation and testing, respectively. The frequency of LiDAR is 20Hz and the annotations are provided at 2Hz. There are 10 classes in total for detection, and among them, 7 moving classes are used for the tracking evaluation. Following the official evaluation protocol, we set the detection and tracking range to [-51.2m, 51.2m]$\times$[-51.2m, 51.2m]. \textbf{Waymo} contains 798 training and 202 validation sequences, and the point clouds are captured by 5 LiDARs at 10 Hz. The official evaluation is carried out in the range [-75m, 75m]$\times$[-75m, 75m], and breaks down the performance on two difficulty levels: LEVEL\_1 and LEVEL\_2, where the former evaluates on objects with more than five points and the latter includes objects with at least one point.

\begin{table*}[h]
\centering
\begin{tabular}{|l|l|cccccccc|}
\hline
            Method & Detection  & Car  & Pedestrian & Bicycle & Motor & Bus  & Trailer & Truck & Overall  \\ \hline \hline
CenterPoint  & Pillar-Det  & 82.5 & \textbf{69.6}       & 20.2 & 40.0  & 78.4 & 40.8  & \textbf{63.9}  & 56.5 \\ \hline
Kalman Filter  & Pillar-Det & 74.7 & 60.3       & 14.0 & 36.0  & 74.9 & 39.1  & 59.5  & 51.2 \\ \hline
 CenterPoint    & Pillar-Track &   79.4   &    64.0        &  24.7    & 53.5      &   \textbf{78.9}   &    46.3   &  59.1     &   58.0   \\ \hline
 Kalman Filter   & Pillar-Track &   76.6 & 68.3 & 25.6 & 54.5 & 74.6 & 45.0 & 57.3 & 57.4\\ \hline
 Ours & Pillar-Track & \textbf{84.1} & 68.3 & \textbf{27.7} & \textbf{57.6} & 76.1 & \textbf{46.6} &  59.2 & \textbf{60.0} \\ \hline \hline 
  CenterPoint &  Voxel-Det & 82.9 & \textbf{73.6} & 40.9 & 54.6 & 79.9 & 48.8 & \textbf{65.2} & 63.7 \\  \hline
 Kalman Filter &  Voxel-Det & 75.7 & 65.7 & 33.5 & 52.2 & 76.7 & 48.2 & 61.1 & 59.0\\ \hline
 CenterPoint  &  Voxel-Track & 81.0 & 70.2 & \textbf{48.0} & 60.6 & 79.7 & 50.9 & 61.1 & 64.5 \\ \hline
 Kalman Filter   & Voxel-Track & 77.5 & 57.3 & 41.5 & 52.4 & 77.2 & 49.4 & 59.1 & 59.2 \\ \hline
 Ours &  Voxel-Track  & \textbf{84.3} & 71.8 & 45.3  & \textbf{64.6} & \textbf{80.5}  & \textbf{54.7} & 61.8 & \textbf{66.1}\\ \hline
\end{tabular}
\vspace{-2pt}
\caption{Comparison of the tracking results using different detection modes with pillar and voxel based backbones on the validation set of nuScenes. We report the overall and per class AMOTA.}
\label{tab:nusc_val}
\end{table*}

\begin{table*}[h]
\centering
\tabcolsep=0.1cm
\begin{tabular}{|l|ccccc|cccc|}
\hline
\multirow{2}{*}{Method} & \multicolumn{5}{c|}{Overall}       & \multicolumn{4}{c|}{Car} \\  \cline{2-10} 
                        & AMOTA$\uparrow$ & FP$\downarrow$ & FN$\downarrow$ & IDS$\downarrow$ & FRAGS$\downarrow$ & AMOTA$\uparrow$   & AMOTP$\downarrow$ & IDS$\downarrow$  & FRAGS$\downarrow$ \\ \hline
AB3DMOT~\cite{weng20203d} &   15.1 & 15088 & 75730 & 9027 & 2557  & 27.8 & 1.325 &  7654 & 1957\\ \hline
Probabilistic-3D~\cite{chiu2020probabilistic}  & 55.0   & 17533 & 33216  & 950 & 776 & 71.9 & 0.580 & 541   & 449\\ \hline
CenterPoint* (Voxel-1440)~\cite{yin2020center} & 63.8 & 18612 & \textbf{22928} & \textbf{760} &529  & 82.9 & 0.384 & 315  & 296 \\ \hline
Ours (Voxel-1024) & \textbf{64.5} &    \textbf{17443} & 26430 & 1042 & \textbf{472} & \textbf{83.6} & \textbf{0.343} & \textbf{214} & \textbf{186} \\ \hline
\end{tabular}
\vspace{-2pt}
\caption{Comparison of the tracking results on the test set of nuScenes. * denotes using deformable convolution and test-time augmentation. Voxel means voxel based backbone, 1024 and 1440 indicate feature map sizes.}
\label{tab:nusc-test}
\end{table*}

\subsection{Evaluation Metrics}
We follow the official evaluation metrics defined by the two benchmarks for comparison. nuScenes adopts the center distance with a threshold of 2m in BEV, i.e., an object within 2m of a ground truth is considered to be true positive. Waymo uses 3D IOU of 0.7 for the vehicle class. It applies MOTA as the main evaluation metric, which
penalizes three error types: false alarms (FP), missing objects (FN) and identity switch (IDS) at each timestamp. Waymo evaluation system automatically picks up a best confidence threshold for MOTA. nuScense on the other hand employs AMOTA to compute the average MOTA under different recalls. We also report FRAGS that counts for the fragments of a track caused by missing detections.

\subsection{Implementation Details}
We implement our approach in PyTorch~\cite{pytorch} based on the codebase of CenterPoint~\cite{yin2020center} and Det3D~\cite{zhu2019class}. We train our models on 8 TITAN RTX GPUs with a batch size of 8 and 4 per GPU for nuScenes and Waymo, respectively. 
Each model is trained for 20 epochs on nuScenes and 12 epochs on Waymo. We utilize AdamW~\cite{adamW} as the optimizer and the one-cycle learning rate scheduling. We apply the standard data augmentation including global rotation and scaling, flipping along X and Y axes, as well as 3D object cut-and-paste from other point clouds.

\begin{table*}[h]
\centering
\begin{subtable}[h]{0.39\textwidth}
\centering
\small
\tabcolsep=0.1cm
\begin{tabular}{|l|cccc|}\hline
                             & AMOTA$\uparrow$                 & AMOTP$\downarrow$                                 & IDS$\downarrow$                   & FRAGS$\downarrow$                  \\ \hline
Unified &  \textbf{60.0} & \textbf{0.774} &  \textbf{1406} & \textbf{412} \\ \hline
Separated     &   44.6    &  0.956   &   4097        &     761   \\ \hline
\end{tabular}
\caption{Comparison of the unified and separated maps on the validation set of nuScenes.}
\label{table:abla-sep}
\end{subtable}
\hfill
\begin{subtable}[h]{0.58\textwidth}
\centering
\small
\tabcolsep=0.12cm
\begin{tabular}{|l|c|ccccc|}
\hline
                 & mAVE$\downarrow$   & Car    & Pedestrian & Bicycle & Motor & Bus    \\ \hline \hline
Baseline (Pillar) & 0.300 & 0.306 & 0.231     & 0.229  & 0.659     & 0.600 \\ \hline
Ours (Pillar)     & \textbf{0.201}   &  \textbf{0.207}      &   \textbf{0.206}         &    \textbf{0.166}    &    \textbf{0.254}        &   \textbf{0.348}     \\ \hline \hline
Baseline (Voxel) &  0.272 & 0.300 & 0.227 & 0.201 & 0.421 & 0.469 \\ \hline
Ours (Voxel) &  \textbf{0.191} & \textbf{0.209} & \textbf{0.208} & \textbf{0.132} & \textbf{0.234} & \textbf{0.304}\\ \hline
\end{tabular}
\caption{Comparison of the velocity estimation on the validation set of nuScenes.}
\label{tab:velo}
\end{subtable}

\vspace{5pt}

\begin{subtable}[h]{\textwidth}
\centering
\begin{tabular}{|c|ccc|ccccccc|}
\hline
\multirow{2}{*}{Combining Maps} & \multicolumn{3}{c|}{Overall} & \multicolumn{7}{c|}{Per Class AMOTA$\uparrow$} \\ \cline{2-11}
&  AMOTA$\uparrow$ & IDS$\downarrow$ & FRAGS$\downarrow$ & Car  & Pedestrian & Bicycle & Motor & Bus & Trailer & Truck  \\ \hline
\xmark & 50.0 & 4761 & 1315 & 79.2 & 51.0 & 14.6 & 33.7 & 71.6 & 45.3 & 54.9 \\ \hline
\cmark &  \textbf{60.0}  & \textbf{1406} & \textbf{412} & \textbf{84.1} & \textbf{68.3} & \textbf{27.7} & \textbf{57.6} & \textbf{76.1} & \textbf{46.6} &  \textbf{59.2} \\  \hline
 \end{tabular}
\caption{Evaluation of combining the current hybrid-centerness map with the previous updated centerness map.}
\label{tab:avg}
\end{subtable}
\label{table:nusc-ablation}

\vspace{5pt}

\begin{subtable}[h]{\textwidth}
\small
\centering
\tabcolsep=0.1cm
\begin{tabular}{|l|cccc|cccc|cccc|}
\hline
\multirow{2}{*}{Resolution} & \multicolumn{4}{c|}{Overall}   & \multicolumn{4}{c|}{Pedestrian} & \multicolumn{4}{c|}{Motor} \\ \cline{2-13} 
                            & AMOTA$\uparrow$ & AMOTP$\downarrow$ & IDS$\downarrow$ & FRAGS$\downarrow$ & AMOTA$\uparrow$  & AMOTP$\downarrow$  & IDS$\downarrow$   & FRAGS$\downarrow$  & AMOTA$\uparrow$  & AMOTP$\downarrow$  & IDS$\downarrow$  & FRAGS$\downarrow$  \\ \hline
1/4 (0.8m)                  &   60.0    &  0.774     &  1406   &   412   & 68.3   & 0.625   & 1216  & 246   & 57.6   & 0.839   & 45   & 9     \\ \hline
1/2 (0.4m)                  & \textbf{61.1}  & \textbf{0.680}  & \textbf{646} & \textbf{320}  & \textbf{75.0}   & \textbf{0.504}   & \textbf{490}   & \textbf{172}   & \textbf{65.9}   & \textbf{0.604}   & \textbf{7}    & \textbf{5}     \\ \hline
\end{tabular}
\caption{Comparison of the tracking results using pillar based backbone with different centerness map resolutions.}
\label{tab:res}
\end{subtable}
\label{table:nusc-ablation}
\vspace{-6pt}
\caption{A set of ablation studies on the validation set of nuScenes.}
\end{table*}

We set the balancing coefficients $(\omega_\text{cen}, \omega_\text{mot}, \omega_\text{reg})$ in Eq.~(\ref{eq:total}) to (1, 1, 0.25) for nuScenes and (1, 1, 1) for Waymo. We extensively compare our approach with Kalman filter and CenterPoint, the two methods that are dominantly used for 3D multi-object tracking. We experiment with two different backbones including pillar based and voxel based to validate the generalizability of our approach. We set pillar size as [0.2m, 0.2m] and voxel size as [0.1m, 0.1m, 0.2m]. Note we do not use higher voxelization resolution for the consideration of computational efficiency during inference.
We follow~\cite{yin2020center} to set the Gaussian heatmap radius when creating the target hybrid-time centerness map.
For thresholding the hybrid-time centerness map, we take the default value 0.1 as the one used for detection in~\cite{yin2020center}. Note this threshold is not further tuned for removing dead tracks so that no additional hyper-parameter is introduced.

\subsection{Results on nuScenes}

\textbf{Validation Set.} Table~\ref{tab:nusc_val} shows the tracking comparisons on the validation set. We report the results in the overall and per class AMOTA. Since a tracking approach is largely affected by the detection performance, for better understanding of SimTrack, we provide two types of tracking performance based on the detection results from (i) Pillar/Voxel-Det: the regularly trained detection models that are used by the original tracking methods; and (ii) Pillar/Voxel-Track: our joint detection and tracking models. 

As demonstrated in Table~\ref{tab:nusc_val}, our approach significantly outperforms the original CenterPoint and Kalman filter by 3.5\% and 8.8\% with the pillar based backbone, and 2.4\% and 7.1\% with the voxel based backbone.
By using the detection results of our approach, the tracking performance of CenterPoint and Kalman filter can be both improved. This is due to our better detection and motion estimation (see details in ablation studies) that benefit from the end-to-end coupled detection and tracking training. Nevertheless, our approach still outperforms both of them by around 2\% when they use our detection results, suggesting that our better tracking performance is not only due to the better detection but also the proposed tracking design. 

It is worth noting that one group of important hyper-parameters for CenterPoint is the maximum distance threshold allowed to be considered matched for each of the different classes. CenterPoint carefully picks the thresholds using the velocity error statistics based on the validation set. Its tracking performance is sensitive to the selected thresholds. For instance, if the threshold for car is changed from 4m to 1m, its AMOTA drops from 82.5\% to 81.0\%, and if the threshold constraint is relaxed to 10m, its AMOTA further drops to 72.1\%. In comparison, our approach fully gets rid of such manually tuned thresholds and is therefore more robust and handy for deployment to new scenarios. 

\textbf{Test Set.} We submit the result of our voxel based model to the testing server of the tracking benchmark of nuScenes. For this submission, we do not use any test-time augmentation. As demonstrated in Table~\ref{tab:nusc-test}, our approach without bells and whistles outperforms the enhanced CenterPoint which is equipped with deformable convolutions and test-time augmentation. Especially for the most important car class in autonomous driving, our model reduces IDS and FRAGS from 315 and 296 to 214 and 186.

\subsection{Ablation Studies}

\begin{figure*}[t]
\begin{center}
\includegraphics[width=0.9\linewidth]{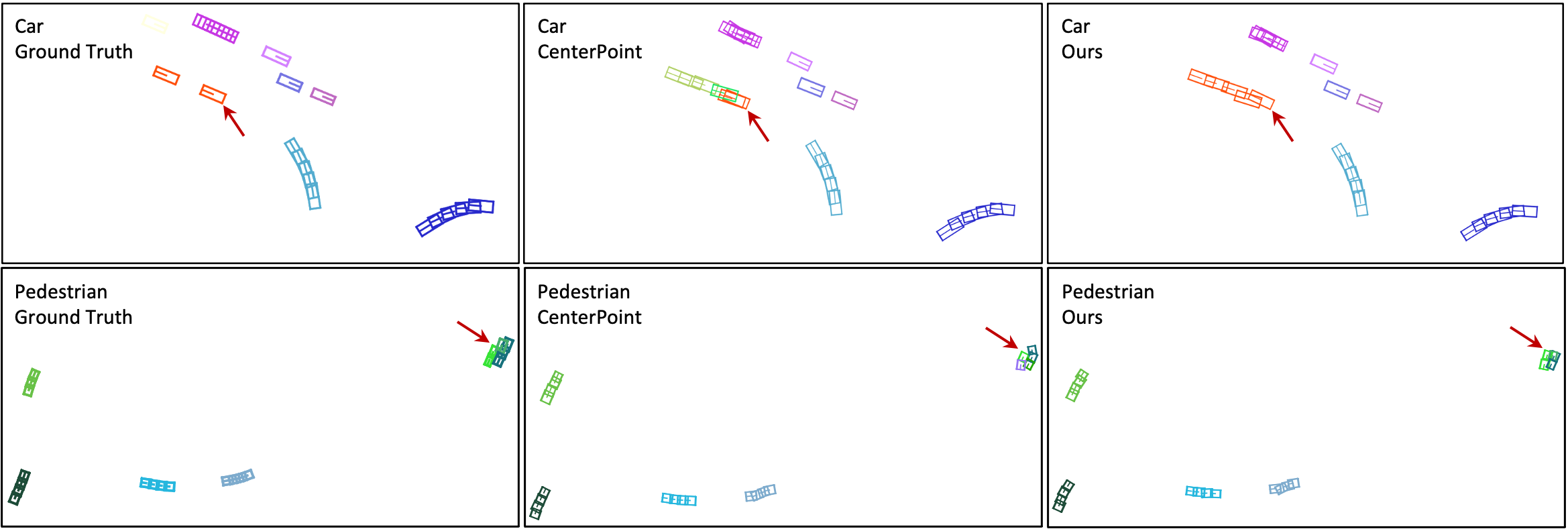}
\end{center}
\vspace{-13pt}
   \caption{Comparison of the qualitative tracking results on the validation set of nuScenes. Each color encodes an object identity over time. Note due to occlusion the ground truth does not provide the annotation of the orange-colored car.}
\label{fig:qualitative}
\end{figure*}

\textbf{Occlusion Analysis.} Being able to handle occlusion is one of the challenges in 3D multi-object tracking as objects can be partially or fully occluded in point clouds for a while. One common practice is to keep dead tracks for a certain number of frames and update their locations by assuming a constant velocity mode. We observe that this heuristic rule substantially impacts IDS, which for example deteriorates from 238 to 500 if dead objects are not retained for certain pre-defined time in CenterPoint. In contrast, SimTrack implicitly tackles occlusion by combining with the confidence scores on previous updated centerness map and updating with the estimated motion. If an object is occluded in current frame but has strong cues in previous frame, our approach is able to keep the object and conjecture its current location. 
Figure~\ref{fig:qualitative} shows one example where the orange-colored car is heavily occluded for a few frames. Our model can successfully keep tracking of this car, while CenterPoint fails to link with the original identity. 
The other example demonstrates that inaccurate velocity estimated by CenterPoint results in identity switch, as velocity estimation for small objects like pedestrian is difficult. Our approach can better handle such cases too. 

\textbf{Unified or Separated Maps.} Here we demonstrate that conducting tracked object association and new-born object detection on a unified map achieves better performance. We also implement a model using separated maps for tracking and detection. Specifically, we modify the hybrid-time centerness map by providing two channels for each class, one for tracked objects and the other for new-born objects. The target assigning strategy remains the same. Table~\ref{table:abla-sep} compares the two designs with the pillar based backbone. It is found that using separated maps performs much worse than using a unified map. We hypothesis that this is due to the extreme unbalancing between the two maps. For a normal scene, the new-born objects only take up a small portion of all objects, thus making training hard. 

\textbf{Combining Maps.} As described in Algorithm~\ref{algo}, the current hybrid-time centerness map is averaged with the previous updated centerness map. In Table~\ref{tab:avg}, we compare this combinatorial design with an alternative that only uses the current hybrid-time centerness map.
The overall and per class results are consistently and considerably improved by combining the two maps. This simple averaging operation provides effective temporal fusion, and in particular, is vital to tackle occlusion as analyzed above.         

\begin{table*}[t]
\centering
\begin{tabular}{|l|ccccc|}
\hline
Method   & MOTA$\uparrow$ & MOTP$\downarrow$ & Miss$\downarrow$ & Miss Match$\downarrow$ & FP$\downarrow$ \\ \hline
Baseline~\cite{sun2020scalability} & 42.5 / 40.1 &  18.6 / 18.6 &  40.0 / 43.4  & \textbf{0.14} / \textbf{0.13}  & 17.3 / 16.4 \\ \hline
  CenterPoint~\cite{yin2020center}     &  51.4 / 47.9 &  17.6 / 17.6 &  47.7 / 41.4 &  0.19 / 0.18 &  \textbf{10.7} / 10.6      \\ \hline
   Ours    &     \textbf{53.1} / \textbf{49.6} & \textbf{17.4} / \textbf{17.4} & \textbf{35.5} / \textbf{39.8} &  0.20 / 0.19 & 11.2 / \textbf{10.5}            \\ \hline
\end{tabular}
\vspace{-4pt}
\caption{Comparison of the vehicle tracking performance on the validation set of Waymo. We report the tracking results with the pillar based backbone, and the numbers are in the format LEVEL\_1 / LEVEL\_2.}
\label{tab:waymo}
\end{table*}

\textbf{Resolution.} Next we show that the performance of our approach can be greatly improved by simply increasing the centerness map resolution, which is in particular effective for small objects such as pedestrian and motorcycle. In the original backbone network, we set pillar size to [0.2m, 0.2m] and downsampling rate to 4, meaning on the centerness map, the size of each cell is [0.8m, 0.8m]. To increase the resolution, we keep the encoder intact but only modify the upsampling layer in decoder to change the downsampling rate to 2. 
Table~\ref{tab:res} shows the overall tracking performance and the specific results of two small object classes: pedestrian and motorcycle. Compared with the lower resolution, using higher resolution improves the tracking performance of the two classes substantially.


\textbf{Velocity Estimation.} In addition to tracking, we show that our approach can produce more accurate velocity for moving objects. We adopt mAVE, the metric officially defined by nuScenes to measure the error of velocity estimation for true positives under different recall rates. 
Table~\ref{tab:velo} reports mAVE of all classes with both pillar based and voxel based backbones. We compare our model with CenterPoint based baseline. For the pillar based backbone, our approach reduces the velocity error by 33\%, in particular for motorcycle, the velocity error is largely reduced by 61\%. It clearly validates that our jointly end-to-end trained detection and tracking model can better exploit the dynamics of moving objects. This improved velocity estimation is potentially beneficial to various downstream tasks such as trajectory prediction and motion planning. 


\subsection{Results on Waymo}
Here we compare the tracking results of the vehicle class on the validation set. In this experiment, we also use the pillar based backbone for the consideration of low latency. As shown in Table~\ref{tab:waymo}, our model delivers clear performance gains over the baseline method provided by Waymo. Compared to CenterPoint, our approach obtains better or on-par results under different metrics. Since nuScenes and Waymo have distinct LiDARs and evaluation metrics, our consistent improvements on the two datasets collectively validate the generalizability of SimTrack. More importantly, we achieve the superior results without requiring the heuristic matching and complex tracking life management as commonly used by the competing algorithms.

\section{Conclusion}
In this paper, we have presented SimTrack, an end-to-end trainable model for 3D multi-object tracking in LiDAR point clouds. Our approach takes a first step to simplify the existing hand-crafted tracking pipelines that involve complex heuristic matching and manual track life management. By combining the proposed hybrid-time centerness map and motion updating branch, our design seamlessly integrates tracked object association, new-born object detection and dead object removal in a single unified model. Extensive experimental results demonstrate the efficacy of our approach. We hope this work can inspire more research toward simple and robust tracking systems for autonomous driving.

\appendix
\section*{Appendix}

In this appendix, Section~\ref{sec:hyper} exemplifies how the representative matching heuristics and related hyper-parameters impact the tracking performance. Section~\ref{sec:more} presents more comparisons between SimTrack and CenterPoint. Section~\ref{sec:speed} reports the inference latency of our model. Section~\ref{sec:waymo} provides more results on nuScenes and Waymo.

\section{Heuristic Matching and Hyper-Parameters}
\label{sec:hyper}
Existing tracking methods involve a number of hyper-parameters in heuristic matching. Some widely used ones include matching threshold, maximum number of frames to keep for a dead track, minimum number of frames before initializing a new track, to name a few. 


It is known that the tracking performance is sensitive to the hyper-parameter setting in heuristic matching.  
For the Kalman filter based tracking, the setting of covariance matrix greatly affects the tracking result. For instance, in~\cite{chiu2020probabilistic} the AMOTA on the validation set of nuScenes is 37.1 when using the default covariance matrix, but the performance boosts to 51.2 after carefully tuning the covariance matrix based on the statistics of prediction errors. 

\begin{table}[b]
\small
\tabcolsep=0.1cm
\begin{tabular}{|l|cc|cccc|}
\hline
Model & Age & Distance & AMOTA$\uparrow$ & AMOTP$\downarrow$ & IDS$\downarrow$ & FRAGS$\downarrow$ \\ \hline
 \multirow{4}{*}{\begin{tabular}[c]{@{}l@{}}Center-\\ Point\end{tabular}} & 3    &  1.0 & 81.0 & 43.3  & 856 & 247 \\ 
  & 3    &  2.0  & 83.1 & 39.5 & 256 & 184 \\ 
 & 3      &    4.0      &    82.5   &  48.0     &  238   & 240      \\ 
 & 3      &     $\infty$       & 59.6      &  49.6    &   318   & 299 \\ 
 &  0     &    4.0      &    80.0   &   48.0    &   365  &     352   \\  \hline \hline
 Ours &  -   &  -      & \textbf{84.1} & \textbf{34.5} & \textbf{148} & \textbf{122} \\ \hline
\end{tabular}
\caption{Impact of the representative heuristics and the setting of related hyper-parameters in the matching step of CenterPoint. We report the tracking results on the validation set (car category) of nuScenes. All results are produced by the pillar based backbone.}
\label{tab:supp-hyper}
\end{table}

To highlight the critical role of setting hyper-parameters for the heuristic matching step, we compare the tracking results of CenterPoint~\cite{yin2020center} with different hyper-parameters in Table~\ref{tab:supp-hyper}. Specifically, we exemplify with two representative hyper-parameters: maximum age and maximum distance. The former is used for a dead track to be retained for a certain number of frames before it is removed. This helps when an object is occasionally occluded in a few frames and shows up again. The latter determines the distance threshold that allows to be matched. CenterPoint tunes this threshold based on the distribution of velocity errors on the validation set. As demonstrated in Table~\ref{tab:supp-hyper}, the two factors significantly impacts the tracking performance. To obtain a reasonably good result, great efforts are in need to tune these hyper-parameters. As a comparison, our approach is heuristic-free but achieves better performance.


\section{More Comparisons with CenterPoint}
\label{sec:more}
\begin{figure*}[t]
\centering

\begin{subfigure}[b]{0.49\textwidth}
\begin{center}

\includegraphics[width=0.95\linewidth]{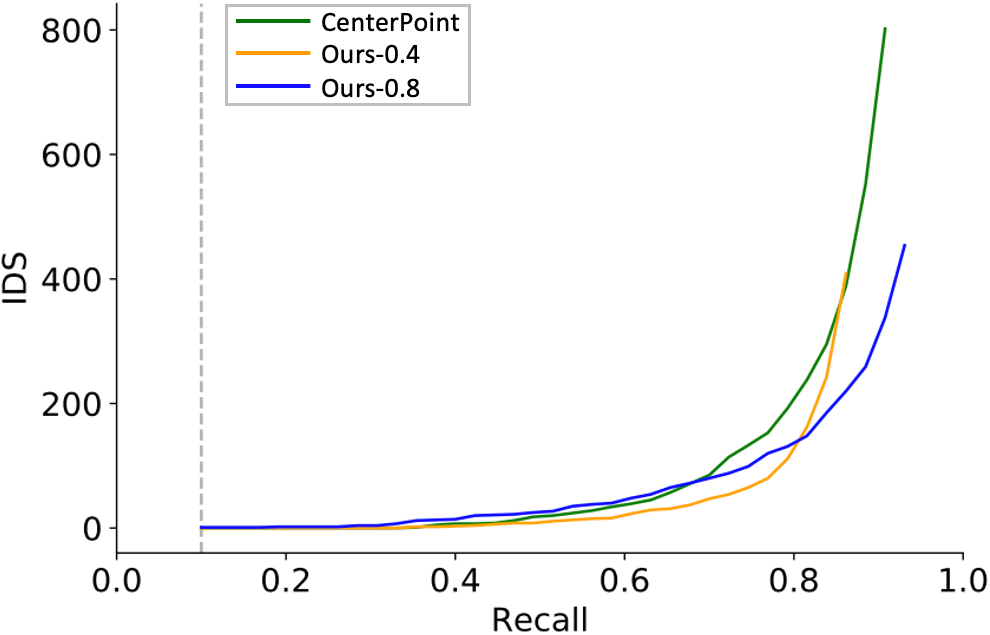}
\end{center}
\vspace{-13pt}
   \caption{IDS-Recall curve of the car category.}
\vspace{8pt}
\label{fig:ids-car}
\end{subfigure}
\hfill
\begin{subfigure}[b]{0.49\textwidth}
\begin{center}

\includegraphics[width=0.95\linewidth]{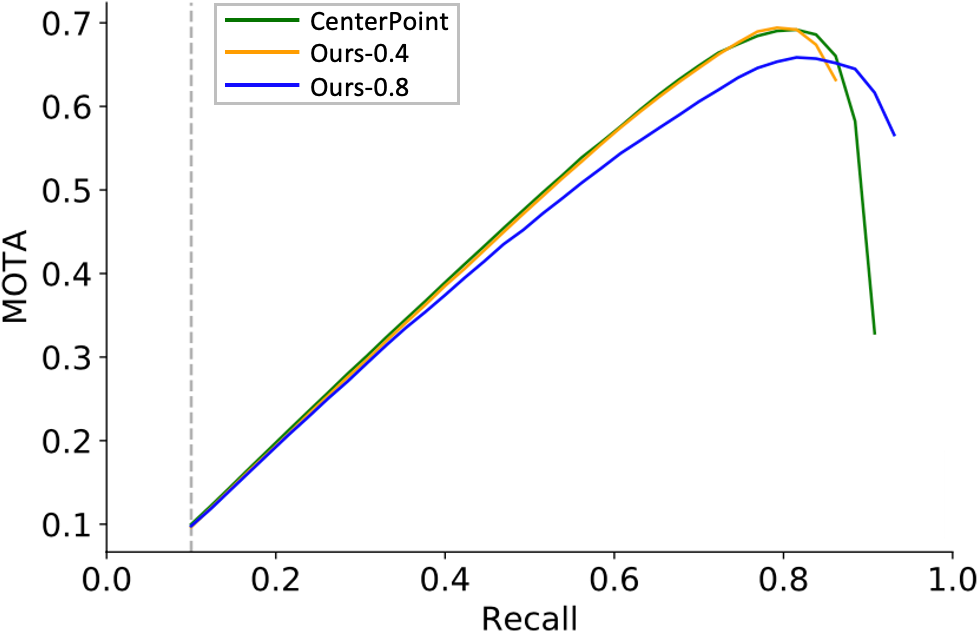}
\end{center}
\vspace{-13pt}
   \caption{MOTA-Recall curve of the car category.}
\vspace{8pt}
\label{fig:mota-car}
\end{subfigure}

\begin{subfigure}[b]{0.49\textwidth}

\includegraphics[width=0.95\linewidth]{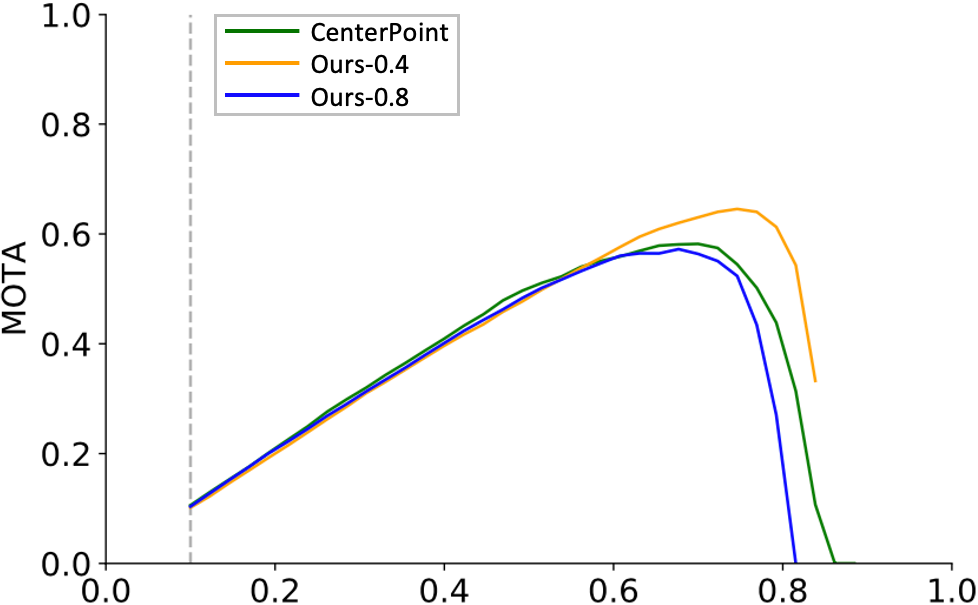}
\vspace{-2pt}
   \caption{MOTA-Recall curve of the pedestrian category.}
\label{fig:mota-ped}
\end{subfigure}
\hfill
\begin{subfigure}[b]{0.49\textwidth}

\includegraphics[width=0.95\linewidth]{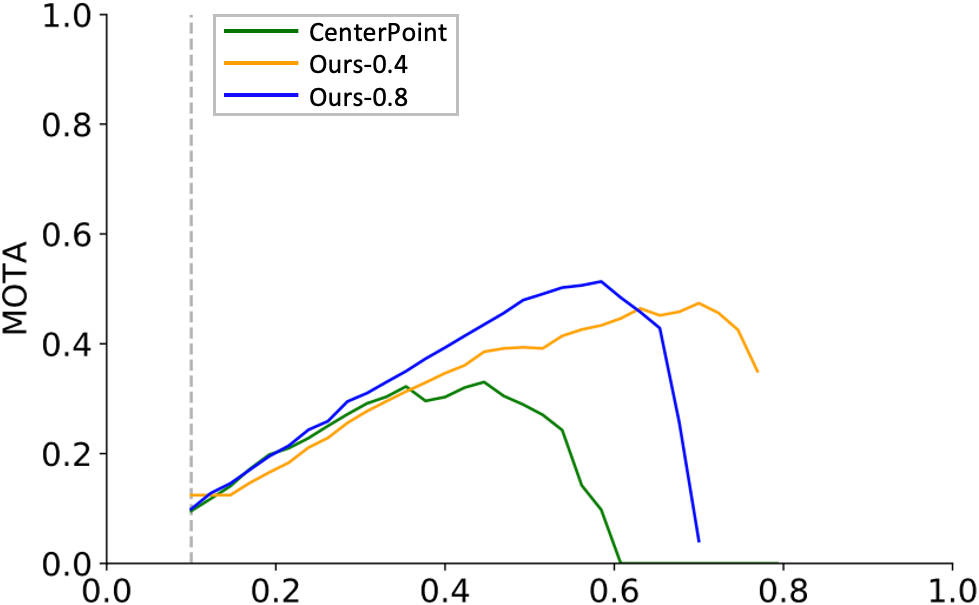}
\vspace{-2pt}
   \caption{MOTA-Recall curve of the motorcycle category.}
\label{fig:mota-motor}
\end{subfigure}
\caption{Comparisons on IDS and MOTA between SimTrack and CenterPoint under different recall rates. All results are produced by the pillar based backbone. Ours-0.4 (0.8) denote the resolution of centerness map: 0.4m$\times$0.4m (0.8m$\times$0.8m). Ours-0.8 is the default resolution. See Section 4.5 in the paper for more details about the resolution. }
\label{fig:supp_comp}
\end{figure*}

Here we provide more detailed comparisons on MOTA and IDS between SimTrack and CenterPoint under different recall rates. As shown in Figure~\ref{fig:ids-car}, our model has much less identity switch under high recall rates. This is because the heuristic matching based tracking methods like CenterPoint suffer from the large amount of false positive detections, while SimTrack is less vulnerable to false positives thanks to our joint detection and tracking design. This advantage makes our approach more robust and stable in particular for the scenarios where a high recall rate is desired. Figures~\ref{fig:mota-car}-\ref{fig:mota-motor} respectively plot the curves of MOTA-Recall for car, pedestrian and motorcycle. Overall, our approach achieves superior MOTA at high recall rates. 


\section{Inference Latency}
\label{sec:speed}
Our joint detection and tracking design is flexible to incorporate in a 3D object detection network and only introduces a small computational overhead to to the backbone network. Table~\ref{time} compares the inference latency between a detection-only model and our joint detection and tracking model using different centerness map resolutions with the pillar and voxel based backbones. As shown in this table, our approach only slightly increases the inference latency of the detection-only model by 1-2ms. We report the inference time on a single TITAN RTX GPU. 


\begin{table}[b]
\centering
\begin{tabular}{|l|c|c|}
\hline
Resolution & Pillar Backbone & Voxel Backbone \\ \hline
0.4m$\times$0.4m  &  36ms / 38ms  &   65ms / 67ms\\ \hline
0.8m$\times$0.8m  &  33ms / 34ms & 63ms / 64ms \\ \hline
\end{tabular}
\caption{Comparison of inference latency of the detection-only model vs. our joint detection and tracking model using different centerness map resolutions and backbones.}
\label{time}
\end{table}

\begin{table}[b]
\centering
\small
\begin{tabular}{|l|cc|cc|cc|}
\hline
\multirow{2}{*}{Method} & \multicolumn{2}{c|}{PointPillars (v)} & \multicolumn{2}{c|}{VoxelNet (v)}  & \multicolumn{2}{c|}{VoxelNet (t)}\\  \cline{2-7}
 & mAP   & NDS  &   mAP & NDS  &   mAP & NDS   \\ \hline
CenterPoint & 50.3 & 60.2  &  56.4 & 64.8   & 58.0 & 65.5\\ \hline
Ours   & \textbf{55.5} & \textbf{64.9}  & \textbf{60.1} & \textbf{67.6} & \textbf{61.3} & \textbf{67.6}\\ \hline 
\end{tabular}
\caption{Comparison of the 3D object detection results on the validation (v) and test (t) sets of nuScenes.}
\label{tab:det}
\end{table}

\begin{table}[t]
\centering
\tabcolsep=0.05cm
\begin{tabular}{|l|cccc|}
\hline
Class   & MOTA$\uparrow$  & Miss$\downarrow$ & Miss Match$\downarrow$ & FP$\downarrow$ \\ \hline
Vehicle  &     54.3 / 50.7 &  34.6 / 38.8 &  0.20 / 0.19 & 10.9 / 10.4    \\ \hline
Pedestrian    &  58.3 / 53.9 &    31.5 / 35.2  &  0.60 / 0.57 &  10.5 / 10.3    \\ \hline
\end{tabular}
\vspace{-4pt}
\caption{We report the tracking performance using dynamic voxelization on the validation set of Waymo, and the numbers are in the format LEVEL\_1 / LEVEL\_2. }
\label{tab:waymo-dynamic}
\end{table}

\section{More Results on nuScenes and Waymo}
\label{sec:waymo}

In addition to simplify and improve tracking, SimTrack can also boost the detection accuracy. Table~\ref{tab:det} compares the detection results of SimTrack and CenterPoint. We report mAP and NDS of all classes on nuScenes. Note the result on the test set of CenterPoint is also based on its enhanced version as described in the paper. Our joint detection and tracking model can significantly improve the detection performance. 
In Table~\ref{tab:waymo-dynamic}, we provide more results of SimTrack on Waymo. We employ the pillar based backbone and adopt the dynamic voxelization proposed in~\cite{zhou2020end} to replace the hard voxelization as used in all other experiments.

{\small
\bibliographystyle{ieee_fullname}
\bibliography{egbib}
}

\end{document}